\documentclass[twocolumn]{article} 
\usepackage[pass]{geometry}
\usepackage{times}
\usepackage{hyperref}
\usepackage{url}
\usepackage{natbib}
\usepackage{amsmath, amssymb}
\usepackage{graphicx}
\usepackage{tabularx}
\usepackage{subfig}
\usepackage[title]{appendix}
\usepackage{amsfonts}
\usepackage{bbm}
\usepackage[color=orange]{todonotes}

\usepackage{tikz}
\usetikzlibrary{automata, positioning}

\graphicspath{ {figures/} }

\DeclareMathOperator{\E}{\mathbb{E}}

\title{Temporal Difference Learning with Neural Networks - Study of the Leakage Propagation Problem}

\author{Hugo Penedones\thanks{equal contribution}, Damien Vincent\footnotemark[1], Hartmut Maennel,\\ Sylvain Gelly, Timothy Mann, Andr\'e Barreto \\
\\
DeepMind and Google Brain - London, UK and Zurich, Switzerland\\
{\small\texttt{\{hugopen, damienv, hartmutm, sylvaingelly, timothymann, andrebarreto\}@google.com}} \\
}

\begin{document}

\maketitle

\begin{abstract}
Temporal-Difference learning (TD) \citep{sutton1988learning} with function approximation can converge to solutions that are worse than those obtained by Monte-Carlo regression, even in the simple case of on-policy evaluation.
To increase our understanding of the problem, we investigate the issue of approximation errors in areas of sharp discontinuities of the value function being further propagated by bootstrap updates. We show empirical evidence of this \textit{leakage propagation}, and show analytically that it must occur, in a simple Markov chain, when function approximation errors are present. For reversible policies, the result can be interpreted as the tension between two terms of the loss function that TD minimises, as recently described by \citep{ollivier2018approximate}. We show that the upper bounds from \citep{tsitsiklis1997analysis} hold, but they do not imply that leakage propagation occurs and under what conditions.
Finally, we test whether the problem could be mitigated with a  better state representation, and whether it can be learned in an unsupervised manner, without rewards or privileged information.

\end{abstract}

\section{Introduction}

Reinforcement Learning \citep{sutton1998introduction, sutton2018introduction} is a framework for studying sequential decision making processes. It deals with the setup where an agent interacts with an environment and at each time step it sees the current \textit{state} (or just an \textit{observation}) and takes one \textit{action}, which will determine the immediate \textit{reward} $R_t$ it will get in the next state.

Typically, the agent is interested in maximising the \textit{expected value} of the return $G_t$, which is often defined as the discounted sum of all rewards in the trajectory.

\[
G_t \doteq R_{t+1} + \gamma R_{t+2} + \gamma^2 R_{t+3} + \dots
\]

A policy $\pi(a|s)$ is a probability distribution over actions, conditioned on the current state. A \textit{state-value function}, $v^{\pi}$, is a function mapping from states to real numbers, defined as the expected return, when the agent starts in a given state $s$ and then behaves according to the policy $\pi$.

\[
v_{\pi} (s) \doteq \mathbb{E}_{\pi}[G_{t}|S_{t}=s]
\]

The environment is formally represented as a Markov Decision Process (MDP), which is a 5-tuple $(\mathcal{S}, \mathcal{A}, P, r, \gamma)$ where $\mathcal{S}$ is the set of states, $\mathcal{A}$ the set of actions, $P(s'|s, a)$ the probability that action $a$ in state $s$ will lead to state $s'$, $r(s, a, s')$ is the reward function, and $\gamma \in [0,1]$ is the discount factor.

\subsection{On-Policy Evaluation}

Estimating the state-value function for the policy that collected the data is the problem known as \textit{on-policy evaluation}, in contrast with the (harder) problem of \textit{off-policy evaluation}, in which one tries to estimate the value function of a different policy \citep{precup2001off, dann2014policy}.
Note that in the \textit{evaluation} or \textit{prediction} problem, one does \textit{not} attempt to find a better or optimal policy, which would be the role of \textit{policy improvement} or \textit{optimisation}. Instead, the goal is just to learn the value function of the current policy as accurately as possible.
In this article we focus on the setup of \textit{offline} or \textit{batch} learning \citep{lange2012batch, riedmiller2005neural}, in which the dataset is pre-collected, rather than a setup where data is being collected \textit{online} simultaneously with the learning process.

Assuming that our value function approximation $\hat{v}_{\pi}$ is parameterised by $\textbf{w}$ (e.g. all neural network weights), and $v_{\pi}$ is the ground truth value function, the goal would be to minimise the Mean-Squared Value Error (MSVE):

\begin{equation}
  MSVE(\textbf{w}) \doteq \lVert v_{\pi}(s) - \hat{v}_{\pi}(s, \textbf{w}) \rVert_{\mu}^{2} \enspace ,
\end{equation}

where $\mu$ is the stationary distribution over states, induced by the policy $\pi$ and the environment dynamics.

Monte-Carlo regression (MC) and Temporal Difference Learning (TD) are the two main algorithms to estimate value functions. The main difference between them is easily visible in their update rules.

Monte-Carlo regression update:

\[
\textbf{w} \leftarrow \textbf{w} + \epsilon [G_t - \hat{v}(S_t, \textbf{w})]\nabla\hat{v}(S_t, \textbf{w})
\]

TD learning update:
\[
\textbf{w} \leftarrow \textbf{w} + \epsilon [R_t + \gamma\hat{v}(S_{t+1}, \textbf{w}) - \hat{v}(S_t, \textbf{w})]\nabla\hat{v}(S_t, \textbf{w})
\]

where $\epsilon$ is the learning rate.

MC uses the actual return $G_t$, available at the end of the trajectory. On the other hand, TD approximates it with the immediate reward plus the discounted estimate of the next state, $R_t + \gamma\hat{v}(S_{t+1}, \textbf{w})$, something that is known as \textit{bootstrapping}.

\subsection{The \textit{leakage propagation} problem}

In the tabular case, where the value function is approximated by a table with one entry per state, Temporal Difference (TD) learning converges to the true value function, see \citep{sutton2018introduction} section 6.3. Not only does it converge to the true value function in the limit, like Monte Carlo regression (MC), it does so at a faster rate. \citet{sutton2018introduction} explain that ``Batch Monte Carlo methods always find the estimates that minimise mean-squared error on the training set, whereas batch TD always finds the estimates that would be exactly correct for the maximum-likelihood model of the Markov process".

However, when combining TD with function approximation, such as the case of neural networks, this no longer holds. Parametric function approximators are typically forced to make approximation trade-offs. In some areas of the state space, there might be sharp discontinuities in the (true) value function, but a neural network function approximator might make a poor fit in those areas, due to shortage of data, or due to limited capacity. Such poor approximations can degrade the accuracy of the value function estimation, in both Monte-Carlo Regression and Temporal-Difference Learning, however with TD Learning the problem is aggravated by further propagating those bad estimates to nearby regions of the state space, after each bootstrap update. We call this problem \textit{leakage propagation}\footnote{Do not confuse the sharp discontinuities in \textit{state space} with the sharp discontinuities of the value function when plotted as a function of \textit{time}. Discontinuities in time are much more natural, and often happen after the collection of a big reward, in an environment where the next reward is far into the future. If the observations immediately before and after the collection of the reward are very different, this wouldn't cause a leakage propagation problem.}.

It is well known that even in the limit the hypothesis that minimises the squared TD error may be worse than the MC solution by a factor of $\frac{1}{1-\gamma}$ where $\gamma \in [0, 1)$ is the discount factor of the MDP \citep{tsitsiklis1997analysis}. A footnote in the same paper even mentions that a much tighter bound, with factor $\frac{1}{\sqrt{1-\gamma^2}}$, can be obtained. These results are upper bounds on the error of the TD fixed point, but we would like to understand better how and under which conditions these errors actually occur. We expand on the relationship with this previous work on section \ref{tsitsiklis_section}.

\subsection{Collecting experimental evidence}

Our first step is to demonstrate the occurrence of \textit{leakage propagation} experimentally. For this, we define a toy environment in which we can expect sharp discontinuities in the value function to happen. In our setup, an agent navigates a two-dimensional space taking fixed-length steps towards some chosen direction. The state is represented by the real-valued $(x, y)$ coordinates. There might be walls, which the agent can not cross, and one or more circle-shaped areas that give a positive reward.

 The task is to estimate the state-value function of a policy in these environments as accurately as possible. Our training dataset is made of 100 independent trajectories, using a random policy initialised at random starting locations. We set the trajectory length limit to 2000 steps, but in the experiments we also show what would happen if the steps limit was lower.
 
\begin{figure}[ht]
\setkeys{Gin}{width=\linewidth}
\begin{tabularx}{\columnwidth}{XX}
\includegraphics{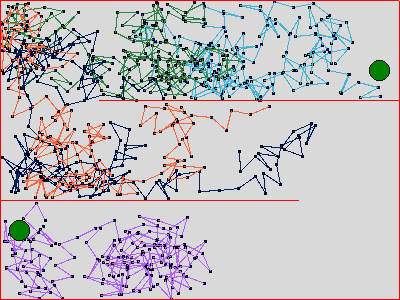} &
\includegraphics{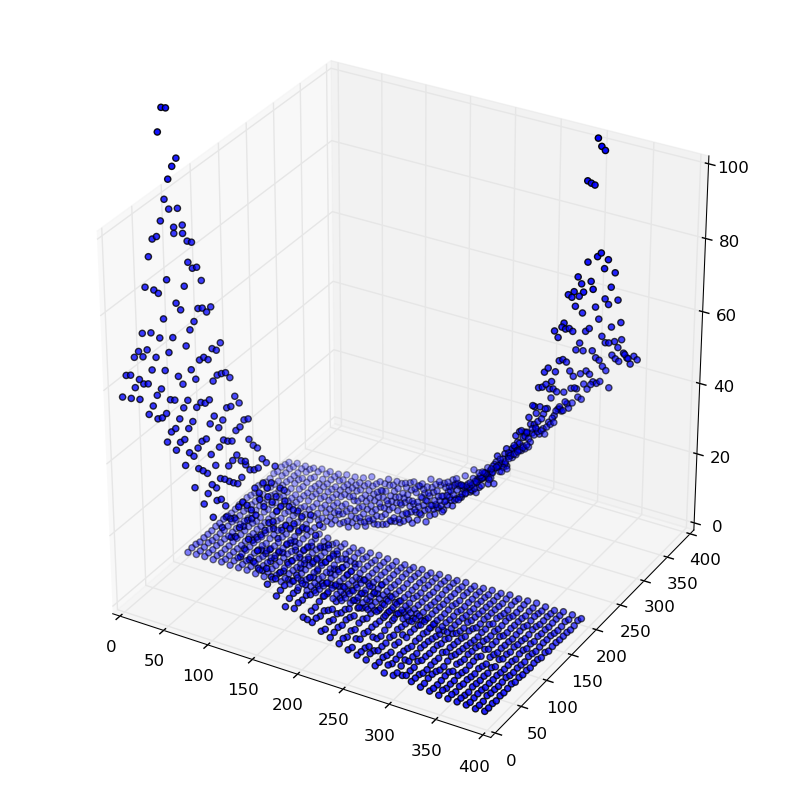} \\
\end{tabularx}
\caption{Random policy trajectories in a S-shaped layout with two reward zones (map 1), and its ground truth value function.}
\label{fig:map_21_layout_and_truth}
\end{figure}

\begin{figure}[ht]
\setkeys{Gin}{width=\linewidth}
\begin{tabularx}{\columnwidth}{XX}
\includegraphics{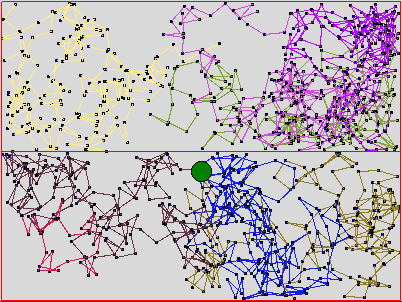} &
\includegraphics{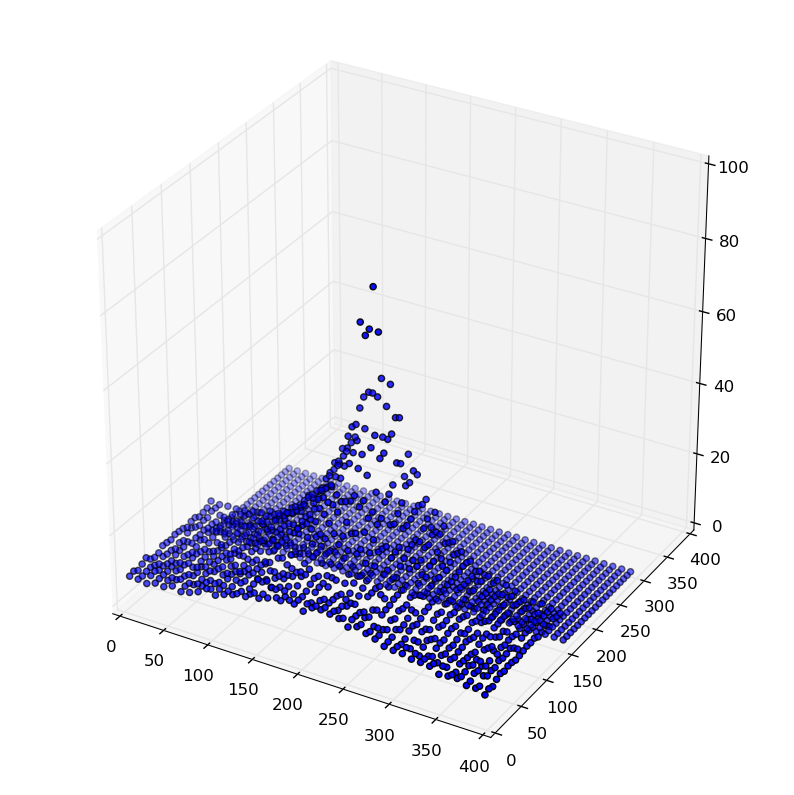} \\
\end{tabularx}
\caption{Random policy trajectories in a layout with two separate rooms (map 2), having a reward zone only in the lower room, and the ground truth value function.}
\label{fig:map_25_layout_and_truth}
\end{figure}

\begin{figure}[ht]
\setkeys{Gin}{width=\linewidth}
\begin{tabularx}{\columnwidth}{XX}
\includegraphics{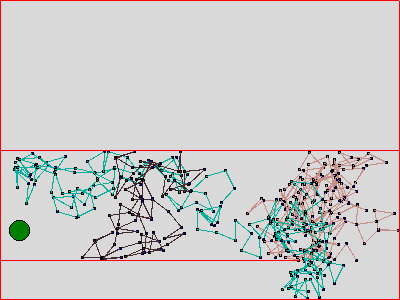} &
\includegraphics{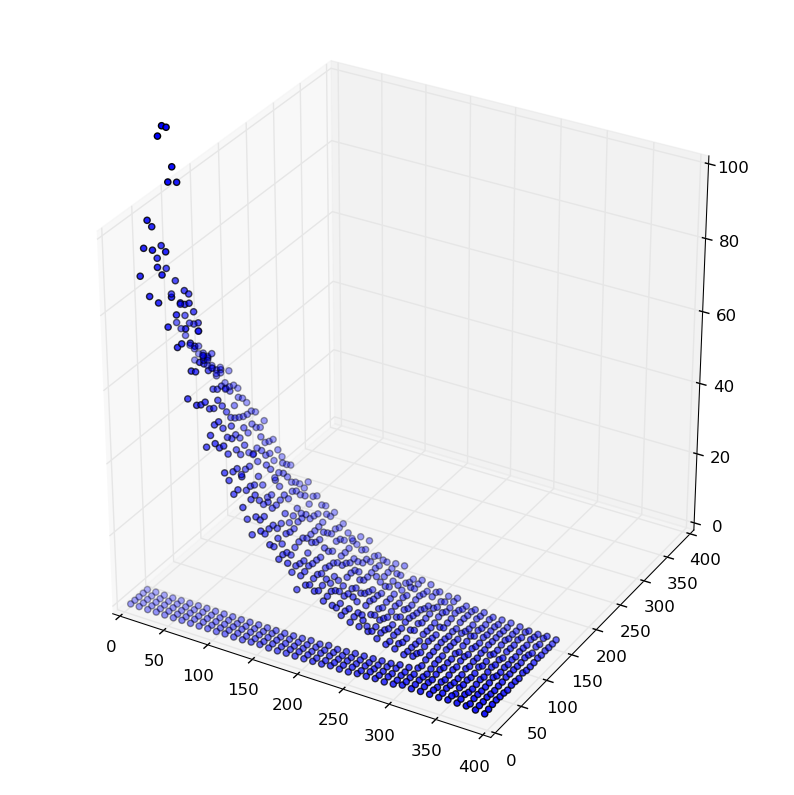} \\
\end{tabularx}
\caption{Left: Random policy trajectories in U-shape layout (map 3). The upper room was not used to generate data (only for padding). Right: the associated ground truth value function.}
\label{fig:map_28_layout_and_truth}
\end{figure}

 For evaluation, we compare the learned value function, with a very precise estimate of the ground truth value function. The ground truth estimate was computed by simulating, in each point of an uniform grid, 1000 trajectories and averaging their Monte-Carlo returns. See Figures (\ref{fig:map_21_layout_and_truth}, \ref{fig:map_25_layout_and_truth} and \ref{fig:map_28_layout_and_truth}) for visualisations of the layouts and the ground truth value functions.

\begin{figure}[p]
\setkeys{Gin}{width=\linewidth}
\begin{tabularx}{\columnwidth}{XX}
\subfloat[MC predictions]{\includegraphics{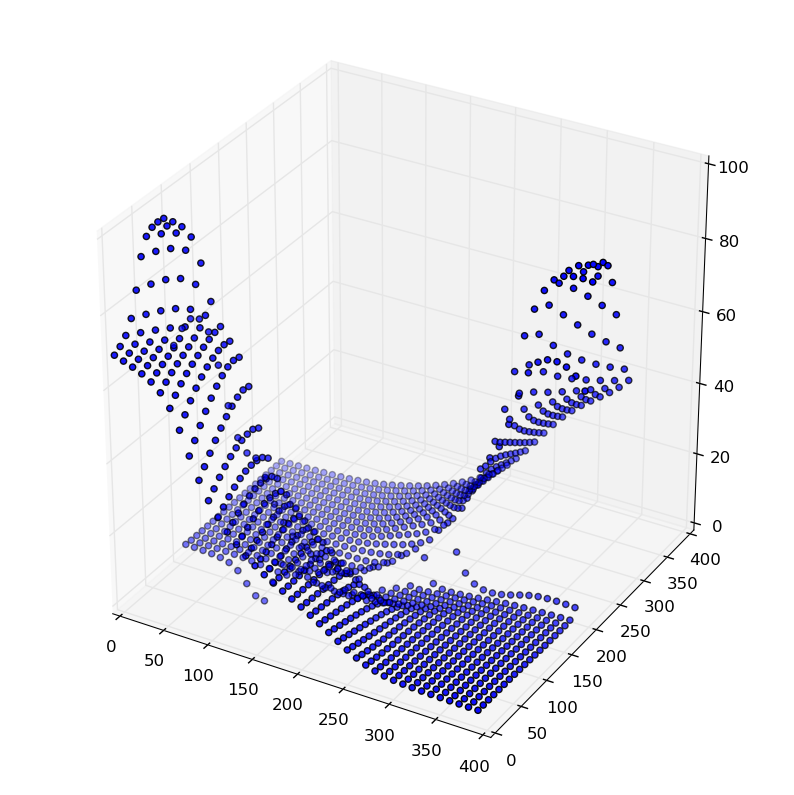}} &
\subfloat[TD predictions]{\includegraphics{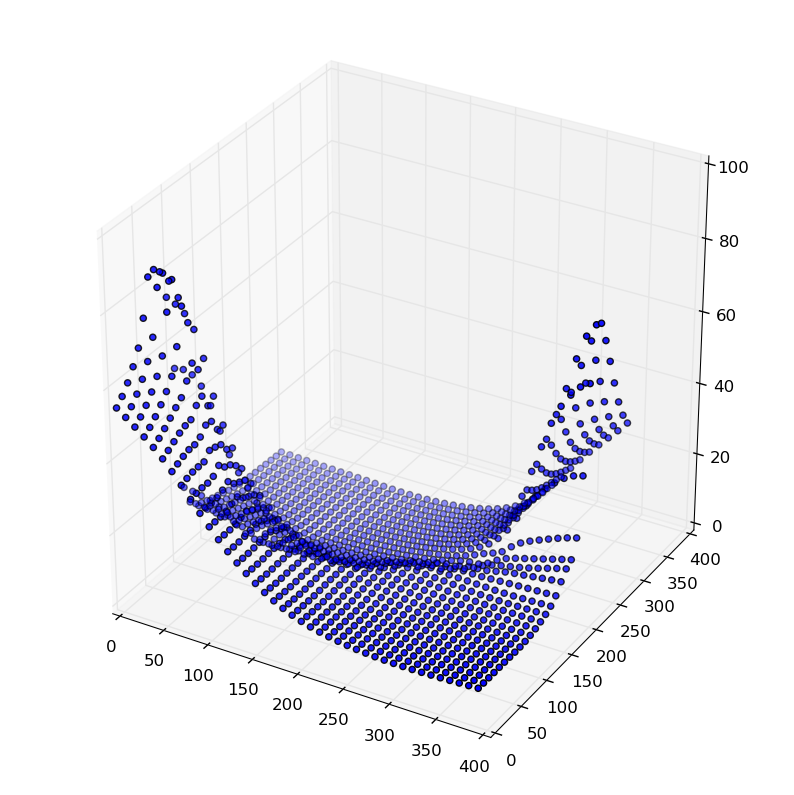}} \\
\subfloat[MC prediction error heatmap]{\includegraphics{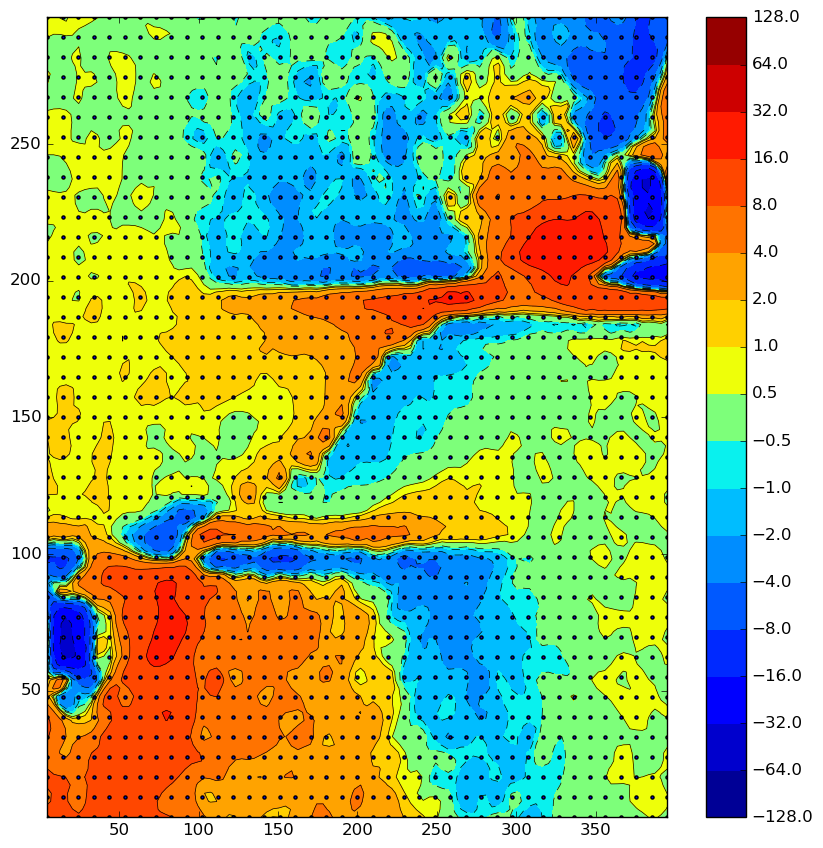}} &
\subfloat[TD prediction error heatmap]{\includegraphics{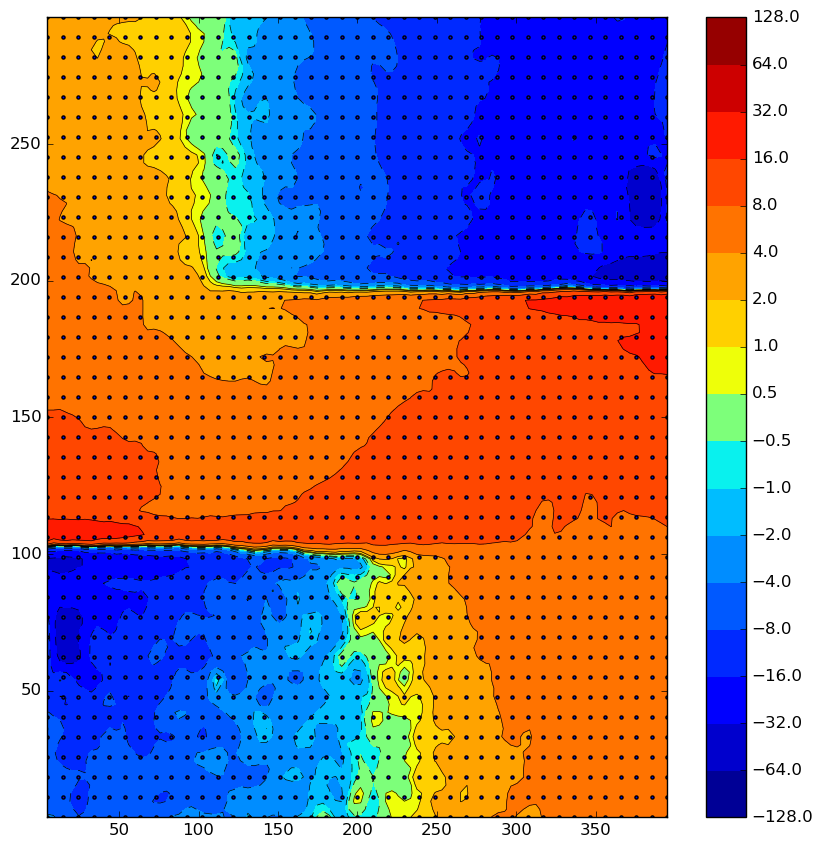}}
\end{tabularx}
\caption{Predictions of MC (left) and TD (right) and error heatmaps compared to ground truth on the S-shaped environment (map 1). Notice that when using TD learning, the middle corridor in the environment, which should have near-zero value, is now being over-estimated, due to leakage propagation from across both walls. }
\label{fig:map_21_td_heatmaps}
\end{figure}

\begin{figure}[p]
\setkeys{Gin}{width=\linewidth}
\begin{tabularx}{\columnwidth}{XX}
\subfloat[MC predictions]{\includegraphics{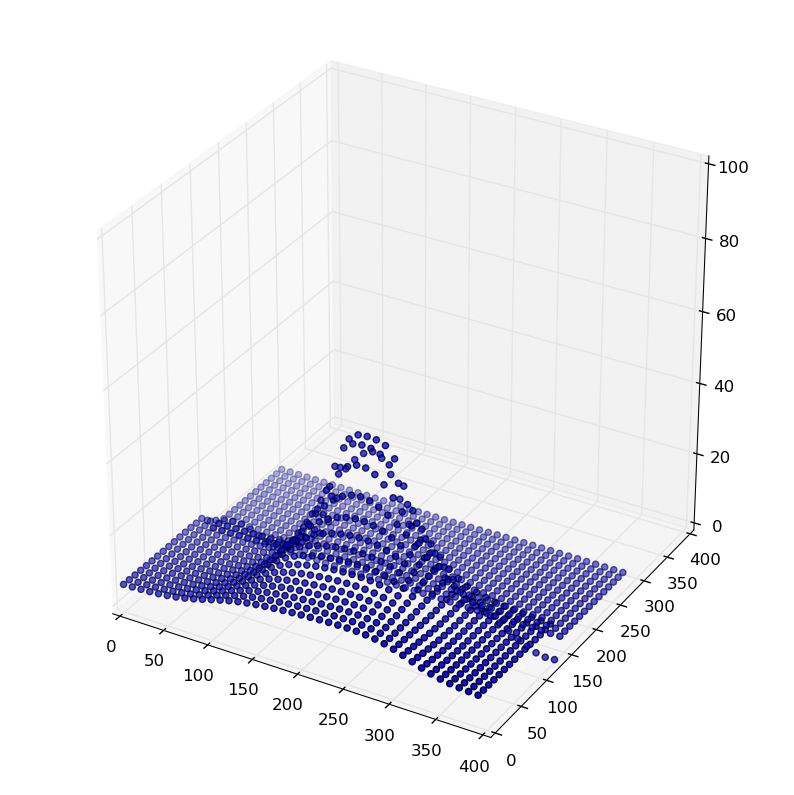}} &
\subfloat[TD predictions]{\includegraphics{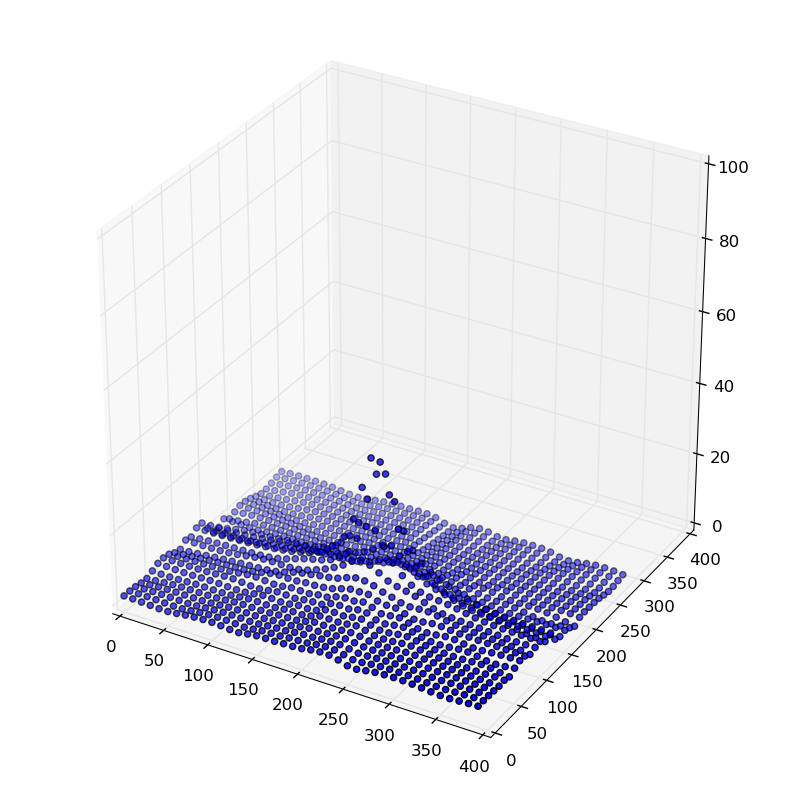}} \\
\subfloat[MC prediction error heatmap]{\includegraphics{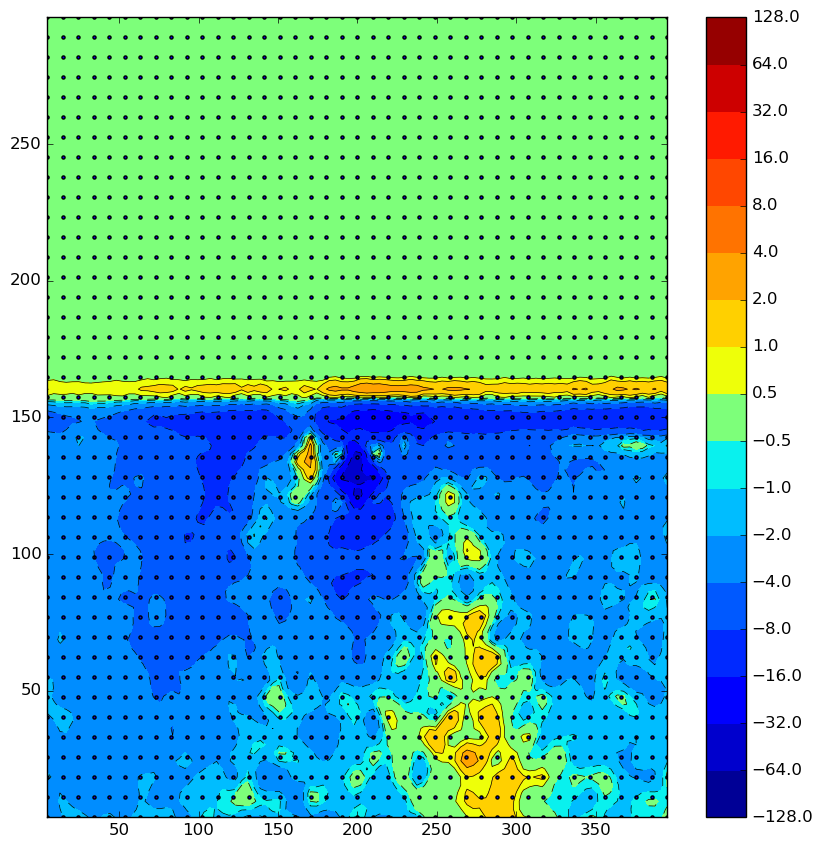}} &
\subfloat[TD prediction error heatmap]{\includegraphics{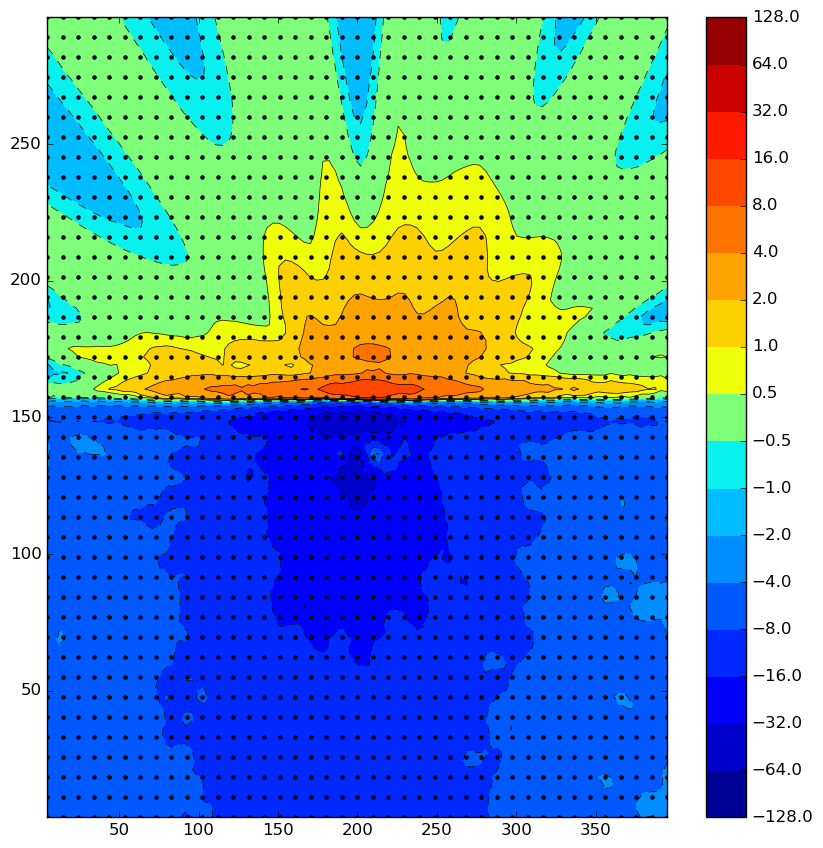}}

\end{tabularx}
\caption{Predictions of MC (left) and TD (right) and error heat-maps compared to ground truth on the environment with two separate rooms (map 2). Notice that when using TD learning, the upper part of the environment, which should have value zero, is now being over-estimated, due to leakage propagation from across the wall. The side that contains the reward zone, also seems under-estimated, suggesting a possible ``negative" leakage from the zero area. In MC the leakage effect is much smaller and it is restricted to regions immediately close to the wall.}
\label{fig:map_25_td_vs_mc_heatmaps}
\end{figure}

As we can see from figures (~\ref{fig:map_21_td_heatmaps} and ~\ref{fig:map_25_td_vs_mc_heatmaps}), MC incurs some approximation error around the walls, but the problem is largely contained in those narrow regions, however in the case of TD, this leakage is propagated much further to other regions.

\section{A more formal understanding}

Now that we have an intuition and demonstrated the problem in practice, we want to understand it more formally. A recent paper by \cite{ollivier2018approximate} describing the loss function that approximate TD minimises can shed some light. The result in that paper is restricted to \textit{reversible} policies, where the probability of a transition and the reverse transition are related by $\mu(s)P(s, s') = \mu(s')P(s', s) \, \forall s, s'$, but it happens to match our navigation task with random policies.

Ollivier shows that if the transition probability $P$ is reversible with respect to its stationary distribution $\mu$, then TD Learning is doing gradient descent over the loss:

\begin{equation}
\gamma \lVert \hat{v} - v \rVert^{2}_{Dir} + (1-\gamma) \lVert \hat{v} - v \rVert^{2}_{\mu}
\label{eq:td_loss_dirichlet}
\end{equation}

where $\gamma$ is the discount factor of the MDP and the $L_2$ norm weighted by the stationary distribution, is just:

\[
\lVert \hat{v} - v \rVert^{2}_{\mu} \doteq \sum_{s} \mu(s) \big[\hat{v}(s) - v(s)\big]^2
\]

The Dirichlet norm $\Vert \hat{v} - v \rVert^{2}_{Dir}$ is defined as:


\[
\frac{1}{2}\sum_{s, s'} \mu(s) P(s, s')\big[(\hat{v}(s') - v(s')) - (\hat{v}(s) - v(s))\big]^2
\]

How does this relate to the leakage propagation problem? Well, note that the second term of the loss in equation (\ref{eq:td_loss_dirichlet}) optimises for the mean square value error that we care about, but the first term optimises for a different norm. In particular, the Dirichlet norm cares about the \textit{differences} in values between consecutive states in our approximation, to be the same as the differences in the true value function. Or equivalently, that the approximation error is the same in the current and next state. This implies that the Dirichlet norm does not distinguish among functions that are shifted by an additive constant. More importantly, as we will see in an example below, under some circumstances the two terms might be under some tension and a solution will compromise between the two, depending on the value of the discount factor $\gamma$.

\subsection{Constraints on sharpness and Leakage Propagation}
As an example, imagine now a scenario in which the state space is the set of non-negative integers $\mathbb{N}$, all transitions go from one integer $s\in\mathbb{N}$ to either $s+1$ or $s-1$, except for $s=0$, in which case the only possible transition goes to $s=1$. Assume further that all rewards are 0, so the true value function is 0 everywhere, and our policy chooses for $s>0$ the transition $s\rightarrow s+1$ with probability $p$ and $s\rightarrow s-1$ with probability $q=1-p$.

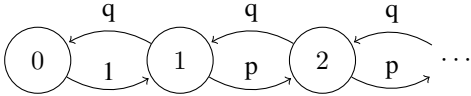
\begin{figure}[ht]
\centering
\begin{tikzpicture}
    \node[state]             (s0) {$0$};
    \node[state, right=of s0] (s1) {$1$};
    \node[state, right=of s1] (s2) {$2$};
    \node[draw=none, right=of s2] (s3) {$\cdots$};

    \draw[every loop]
        (s0) edge[bend right, auto=left]  node {1} (s1)
        (s1) edge[bend right, auto=right] node {q} (s0)
        (s1) edge[bend right, auto=left]  node {p} (s2)
        (s2) edge[bend right, auto=right] node {q} (s1)
        (s2) edge[bend right, auto=left]  node {p} (s3)
        (s3) edge[bend right, auto=right] node {q} (s2);        
\end{tikzpicture}
\caption{Simple infinite Markov chain with transition probabilities p and q.} \label{fig:markov_chain}
\end{figure}

We will assume that $0 < p < 0.5$, since for $p\geq 0.5$ we would not have a stationary distribution, the process would ``wander off to infinity''.
For $p < 0.5$ the stationary distribution $\mu$ on $S=\mathbb{N}$ has probabilities proportional to
\begin{equation}
    q, 1, \frac{p}{q}, \frac{p^2}{q^2}, \frac{p^3}{q^3}, ...
    \label{eq:stable_dist}
\end{equation}

We assume also that the class of permissible approximations has as (only) condition that $v(0) = \alpha$ for some $\alpha>0$. This is a very idealised case of a wall like in Figure (\ref{fig:map_25_layout_and_truth}) - imagine there are also states for integers $s < 0$, and they have positive value functions, but they cannot be reached from $s\geq 0$, and our function class somehow fixes the value at $s=0$ to $v(0)=\alpha$ when the correct values at $s< 0$ are given.

Assuming we start with some estimate for our value function $v:\mathbb{N}\rightarrow\mathbb{R}$, what is the solution to which TD will converge?
Without restrictions, there are two possibilities, however we will assume that we start
from a bounded estimate (it would also be enough to assume $v\in L^2(\mathbb{N}, \mu)$, which is a weaker condition).

We will first analyse this directly, and later see how this can also interpreted as minimum of (\ref{eq:td_loss_dirichlet}).
For compactness we abbreviate $v(s)$ as $v_s$.

The average TD update at a state $s>0$ with learning rate $\epsilon$ would be
\begin{equation}
  v_s \leftarrow v_s +  \epsilon\left(p\gamma \cdot v_{s+1} + q \gamma\cdot v_{s-1} - v_s\right)
  \label{eq:TD_update_direct}
\end{equation}
By rewriting the right hand side as
\begin{equation*}
  (1-\epsilon) \cdot v_s +  \epsilon\cdot \gamma(p\cdot v_{s+1} + q\cdot v_{s-1})
\end{equation*}
we see that if $v$ is bounded by $|v(s)|<B$ for some $B\in\mathbb{R}$, the same is true for the 
updated value function. So the updates can only converge to a bounded solution (and it is also true that the $L_2(\mathbb{N},\mu)$ norm does not increase, which we would use if we only assume
we started from a $v\in L_2(\mathbb{N},\mu)$.)

\begin{figure}[ht]
\begin{center}
\includegraphics[scale=0.4]{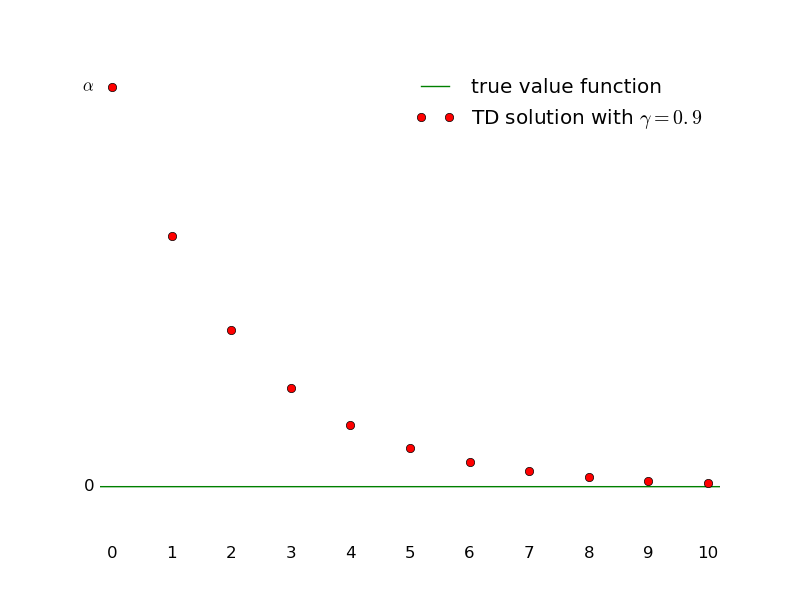}
\end{center}
\caption{Plot of the analytic solution for TD in a simple discrete 1-d problem. Assuming the true value function (green) is zero, but we are forced to make an approximation mistake in state $s=0$, by setting its value to $\alpha>0$. After that mistake is incurred, TD will not make a sharp correction to the value function, but rather decay it exponentially towards zero (red). This formally demonstrates the leakage propagation effect with approximate TD.}
\label{fig:td_exp_decay_solution}
\end{figure}

The update \eqref{eq:TD_update_direct} leaves $v_s$ for the state $s>0$ invariant iff
\begin{equation}
v_{s+1} - \frac{1}{p\gamma}v_{s} + \frac{q}{p}\cdot v_{s-1} = 0
\label{eq:values_recurrence}
\end{equation}
Finding the roots of the characteristic equation:
\begin{equation}
r^2  - \frac{1}{p\gamma}r + \frac{q}{p} = 0
\label{eq:char_eq}
\end{equation}
we get the two solutions:
\begin{equation*}
   r_{1,2} = \frac{1\pm \sqrt{ 1 - 4pq\gamma^2}}{2p\gamma}
\end{equation*}
which are real and positive for any $\gamma < 1$ since $4pq<1$. Let $r_1$ be the smaller one of them. Since
$\gamma = (p+q) \cdot \gamma < 1$, we have $q\gamma < 1-p\gamma$ and
\[
    1-4pq\gamma^2 > 1-4q\gamma(1-p\gamma) = (1-2p\gamma)^2
\]
from which it follows that
\[
   1 - \sqrt{1-4pq\gamma^2} < 2p\gamma
\]
and hence $r_1<1$. This computation also shows that
\begin{equation}
   r_1\rightarrow 1 \qquad \hbox{for} \ \gamma\rightarrow 1
   \label{eq:limit1}
\end{equation}
Since \eqref{eq:char_eq} shows that $r_1\cdot r_2 = q/p$, we have
\[
  0 < r_1 < 1 < q/p < r_2
\]
We can write the general solution for the recurrence as:
\[
   v_s = c_1 r_1^s + c_2 r_2^s
\]
Since $v$ is bounded (in fact, $v \in L^2(\mathbb{N},\mu)$ is enough), we must have $c_2 = 0$. On the other hand $s=0$ gives $c_1+c_2 = \alpha$, and therefore:
\[
    v_s = \alpha r_1^s  \qquad \hbox{for}\ \ s=0,1,2,...
\]
In other words, in the optimal solution, the value function $v$ that TD finds decays exponentially, as states move away from the constrained state, see Figure (\ref{fig:td_exp_decay_solution}).

Because of \eqref{eq:limit1}, the closer the discount factor $\gamma$ is to $1$, the slower the decay will be, which means more leakage propagation. This makes intuitive sense, because it means we are bootstrapping more heavily. If $\gamma \rightarrow 1$ in fact, we get that $v_s \rightarrow \alpha$, which is a totally flat, grossly overestimating region.
When $\gamma\rightarrow 0$, $r_1\approx q\gamma \rightarrow 0$ and therefore $v_s \rightarrow 0$ for $s>0$, which means that there is no leakage propagation and the mean-squared error is minimal. This corresponds to the non-interesting situation of just predicting immediate rewards, however.

\subsection{Relation to Ollivier [2018]}
We now relate this to \eqref{eq:td_loss_dirichlet}.
As we can see directly from \eqref{eq:stable_dist}, our simple example satisfies the reversibility condition, as $\mu(s_t)P(s_t, s_{t+1}) = c\frac{p^{t}}{q^{t-1}} = \mu(s_{t+1})P( s_{t+1}, s_t)$, where $c$ is the constant needed to get the stationary distribution probabilities to sum to 1.
Since the true value function in our example is $v_t=0$ for all $t$ and noting that $p\mu(s_{t-1})=q\mu(s_{t})$ for $t>1$, we sum all over transitions to get the Dirichlet term, $\lVert \tilde{v} - v \rVert^{2}_{Dir}$, of the loss function \eqref{eq:td_loss_dirichlet}:
\[
\mu(0)\alpha^2 + \sum_{t=1}^{\infty} \frac{1}{2} \big[p\mu(t)(\tilde v_{t+1} - \tilde v_{t})^2 + q\mu(t)(\tilde v_{t} - \tilde v_{t-1})^2]
\]

while the $L_2$ term $\lVert \tilde{v} - v \rVert^{2}_{\mu}$ is just $\sum_{t=0}^{\infty} \mu(s_t)\tilde v_{t}^2$. If we solve for $\frac{\partial}{\partial \tilde v_i}(\gamma \lVert \tilde{v} - v \rVert^{2}_{Dir} + (1-\gamma) \lVert \tilde{v} - v \rVert^{2}_{\mu}
) = 0$ to find its minima, we recover the recurrence previously mentioned in \eqref{eq:values_recurrence}. 

In the paper \cite{ollivier2018approximate} a finite state space was assumed, but the
computation of the gradient of \eqref{eq:td_loss_dirichlet} is a purely algebraic transformation, so
it is still valid for our case if everything converges.

Now we can understand the leakage as a consequence of the tension between the two components of the loss: according to the $L_2$ norm one would set all $v_s = 0$ for $s\neq 0$, achieving the minimal first part of the cost $\gamma\cdot|v|_\mu^2 = \gamma\cdot\alpha^2$. However, according to the second part $(1-\gamma)\cdot|v|_{Dir}^2$, the best would be to actually set all $v_s = \alpha$, assuring that the estimates are ``as flat" as the true value function in that region, and incurring zero cost for the Dirichlet norm part. Given that the full loss is a mixture of the two convex loss functions which have no minimum in common, neither of the minima of these two components can be a minimum for the mixture, so in particular, leakage must occur.


\subsection{Relation to Tsitsiklis and Van Roy [1997] } \label{tsitsiklis_section}
We can look at this example also from the point of view of \cite{tsitsiklis1997analysis}.
This paper defines
operators $T^{(0)}$ and $\Pi$ on $L^2(S, \mu)$, where $\mu$ is the stable distribution on $S$,
$\Pi$ is the orthogonal projection on the function class, and $T^{(0)}$ is the average TD(0)
update (see below).

Our case does not directly satisfy the assumptions in \cite{tsitsiklis1997analysis}, since our
function space is not a vector space, but an affine space: For
\[
    a \doteq (\alpha, 0, 0, ...)
\]
we can write it as $a+W$ with the vector space
\[
   W \doteq \left \{ (v_0, v_1,...)\in L^2(\mathbb{N}, \mu)\,|\, v_0 = 0  \right\}
\]
We will consider both the orthogonal projection $\Pi$ onto $a+W$ and the orthogonal
projection $\Pi_0$ onto $W$:
\begin{eqnarray*}
  \Pi_0 (v_0, v_1,...) & \doteq & (0, v_1,v_2,...) \\
  \Pi (v_0, v_1,...) & \doteq & (\alpha, v_1,v_2,...) 
\end{eqnarray*}

Since we do not consider other $T^{(\lambda)}$ , we will just write $T$ for the operator $T^{(0)}$ on $L^2(S,\mu)$, it is given by
\[
    Tv(s) \doteq \begin{cases}
           \gamma\cdot v(1) & \hbox{for}\ s=0\\
           \gamma\cdot \big(p\cdot v(s+1) + q\cdot v(s-1)\big)  &\hbox{for}\ s>0
       \end{cases}
\]
So our update \eqref{eq:TD_update_direct} leaves $\tilde v$ invariant if and only if
it is a fixed point of $\Pi T$, as in \cite{tsitsiklis1997analysis}. On the other hand
the ideal solution $v^*=(0,0,...)$ is a fixed point of $T$.

To get the
``$1/(1-\gamma)$ bound'' of this paper, we adjust their computation to our situation:
\begin{eqnarray}
   |\tilde v - v^*| &\leq& |\tilde v - \Pi v^*| + |\Pi v^* - v^*| \label{eq:triangle_ineq}\\
     &=& |\Pi T \tilde v - \Pi T v^*| + |\Pi v^* - v^*| \nonumber \\
     &=& |\Pi_0 T (\tilde v - v^*)| + |\Pi v^* - v^*| \nonumber
\end{eqnarray}
The proof of Lemma 1 (p. 680) in \cite{tsitsiklis1997analysis}
then shows that $|Tv| \leq \gamma |v|$ for
all $v\in L^2(S,\mu)$, and since obviously $|\Pi_0v|\leq |v|$, also $\Pi_0 T$ is a contraction with factor $\leq\gamma$, from which we get
\begin{equation}
   |\tilde v - v^*| \leq  \frac{1}{1-\gamma} |\Pi v^* - v^*|
   \label{eq:Tsitsiklis}
\end{equation}
In fact, we can replace \eqref{eq:triangle_ineq} by the equality
\[
     |\tilde v - v^*|^2 = |\tilde v - \Pi v^*|^2 + |\Pi v^* - v^*|^2
\]
because $\tilde v - \Pi v^* \in W$ and $\Pi v^* - v^*$ is orthogonal to $W$. 
This gives (in otherwise the same way) the sharper bound
\begin{equation}
   |\tilde v - v^*| \leq  \frac{1}{\sqrt{1-\gamma^2}} |\Pi v^* - v^*|
   \label{eq:Tsitsiklis2}
\end{equation}
which is mentioned in the footnote of \cite{tsitsiklis1997analysis}.

The above proofs used the facts
\begin{itemize}
    \item $T v^* = v^*$
    \item $\Pi T \tilde v = \tilde v$
    \item $\Pi_0 T$ is contracting with factor $\gamma$
\end{itemize}
about $T,\tilde v, v^*$.
To see whether the factor $1/\sqrt{1-\gamma^2}$ that occurred in this proof
is related to our ``leakage observation'', consider another operator / estimate pair
\begin{eqnarray*}
   T'v(s) &\doteq& \gamma\cdot v(s) \\
   \tilde v' &\doteq& (\alpha, 0, 0, 0,...)
\end{eqnarray*}
which has the same properties (even though $\tilde v'$ is without ``leakage'').
The proof gives the same bound, but in fact we even have
\begin{equation}
   |\tilde v' - v^*| =  |\Pi v^* - v^*|
   \label{eq:counterex}
\end{equation}
The reason that $1/\sqrt{1-\gamma^2}$ appeared in \eqref{eq:Tsitsiklis2} 
was that we only know that
$\Pi_0 T$ is contracting with factor $\gamma$, so without knowledge about $\tilde v$
we only can say that
\[
   |\Pi_0 T (\tilde v - v^*)| \leq \gamma \cdot |\tilde v - v^*|
\]
from which \eqref{eq:Tsitsiklis2} followed.
However, to prove that the constant in \eqref{eq:Tsitsiklis2} cannot be lowered
from $1/\sqrt{1-\gamma^2}$ to 1 (as in \eqref{eq:counterex}), we would need to give a lower bound to
$|\Pi_0 T (\tilde v - v^*)|$. In the $T', \tilde v'$ example ``without leakage'' this expression is 0;
if we do have leakage (i.e. values $\tilde v_s>0$ for $s>0$) in $\tilde v$, then this expression is
$>0$.
So in a sense the possibility of leakage is indeed responsible for a factor $>1$ in
\eqref{eq:Tsitsiklis2} in this example, but whether we have leakage or not does not seem to follow
from the general arguments in the above proof.

\section{Better state representations}

We saw that the existence of sharp discontinuities in the state-value function causes TD to miss-behave in combination with function approximation. In this section, we investigate what would happen if we had a better state representation and whether we can hope to learn it from the data, without privileged information and without using the reward signal.


We start by noting that if two states in our trajectories have very similar input features, but it takes a large number of steps to connect them, following the policy, we do not want the function approximator to extrapolate from one to the other, by continuity or smoothness. Instead, we want these two states to be pushed apart, in some initial embedding space, so that the na\"ive extrapolation does not happen by default. However, if two states occur just after each other in the same trajectory, we assume it is safer to keep them close to each other in the embedding space. Even if they have a big intermediate reward in between, there is already enough incentive from the TD error to differentiate between the two.

In order to compare different state representations, hand-designed or learned, we start by defining a two stage architecture, composed of an embedding network followed by a value network.

\subsection{Embedding and value networks}

Our baseline neural network is a Multi-Layer Perceptron (MLP) which takes as input the state representation $s$ and outputs and single scalar $\hat{v}(s, \textbf{w})$. To experiment with different representations, keeping the model capacity constant, we split the MLP into two sets of layers: the embedding layers, with weights $\textbf{w}_e$ and the value layers, with weights $\textbf{w}_v$. The final output is therefore a composition of the two functions: $\hat{v}(s, \textbf{w}) \equiv g( f(s, \textbf{w}_e), \textbf{w}_v)$.

The training procedure is now divided into two stages: first learn the embedding network $f$ and freeze its weights $\textbf{w}_e$; and \textit{then} learn the top network layers $g$ to approximate the value function.

In the next subsections we suggest some possible directions to create such representations. We start by illustrating the opportunity gap with a hand-designed transformation and then try out two unsupervised ways to learn the intermediate representation from data.

\subsection{Hand-designed Oracle Embedding}

As a simple sanity check, we hand-craft a transformation that separates nearby points across the walls perfectly, assuming privileged knowledge about the map layout. If this does not work, there is little hope for learning such representation without supervision.

\begin{figure}[ht]
\setkeys{Gin}{width=\linewidth}
\begin{tabularx}{\columnwidth}{XX}
\subfloat[$\alpha=0.0$]{\includegraphics{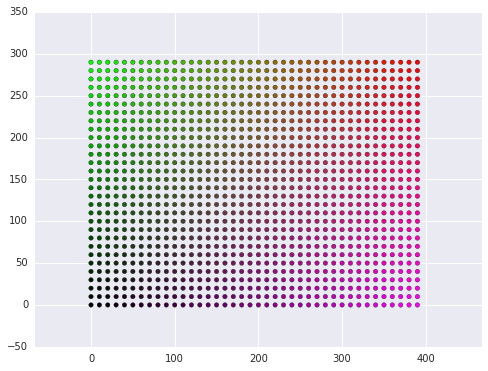}} &
\subfloat[$\alpha=0.25$]{\includegraphics{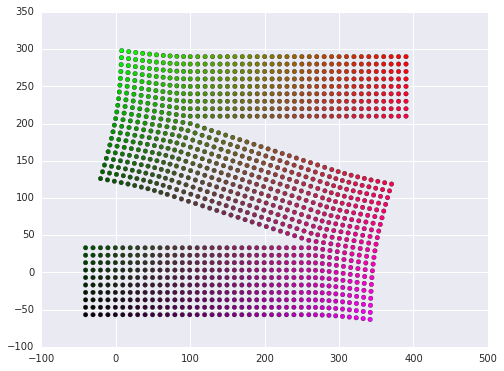}} \\
\subfloat[$\alpha=0.5$]{\includegraphics{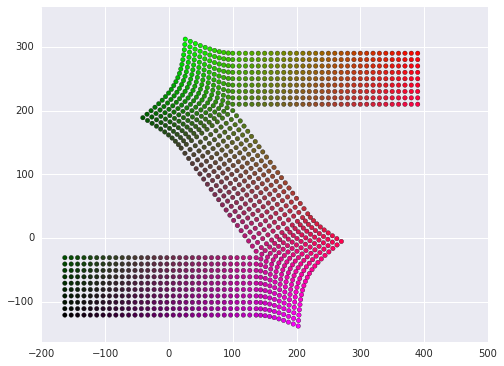}} &
\subfloat[$\alpha=1.0$]{\includegraphics{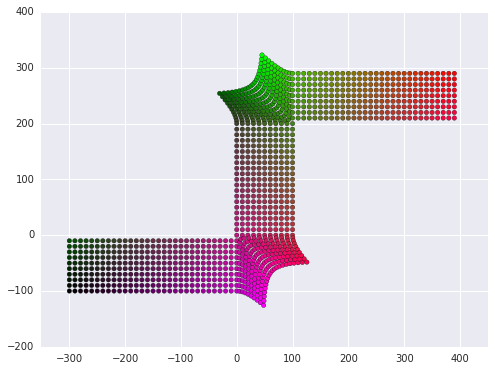}}
\end{tabularx}
\caption{Hand-designed ``oracle" embedding with different degrees of separation across the labyrinth walls for the S-shaped layout (map 1). This assumes privileged knowledge about the exact location of the walls. The amount of separation of the independent ``corridors" in this layout is controlled by varying the $\alpha$ parameter from 0 to 1.}
\label{fig:oracle_embedding_angles}
\end{figure}

\begin{figure}[ht]
\setkeys{Gin}{width=\linewidth}
\begin{tabularx}{\columnwidth}{XXX}
	\includegraphics{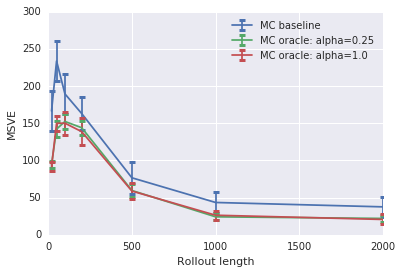} &
	\includegraphics{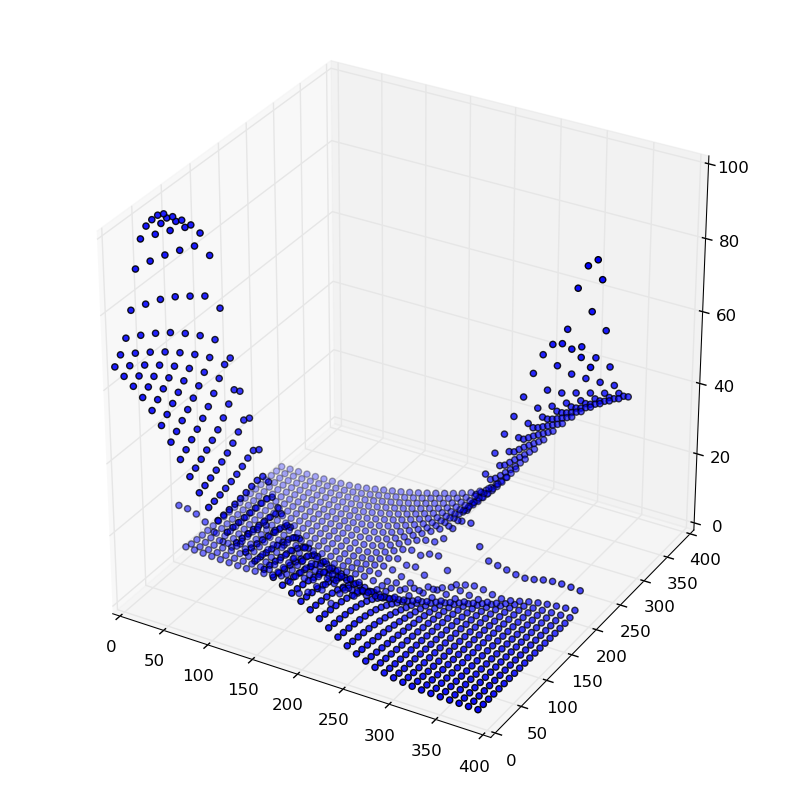} &
	\includegraphics{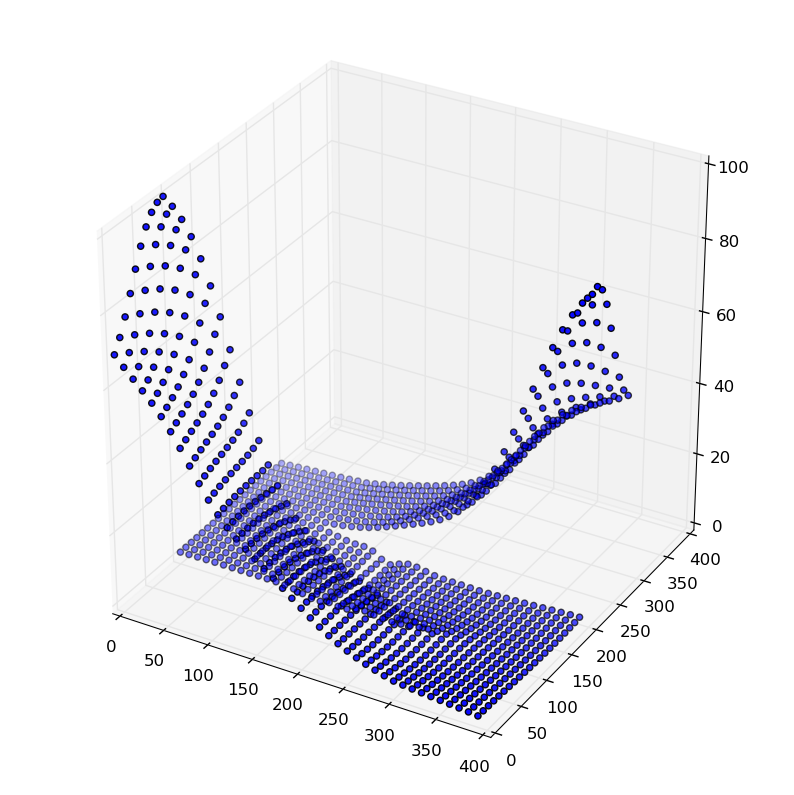} \\
	\includegraphics{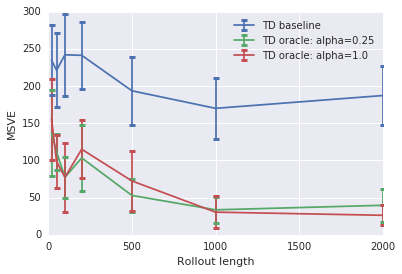} &
	\includegraphics{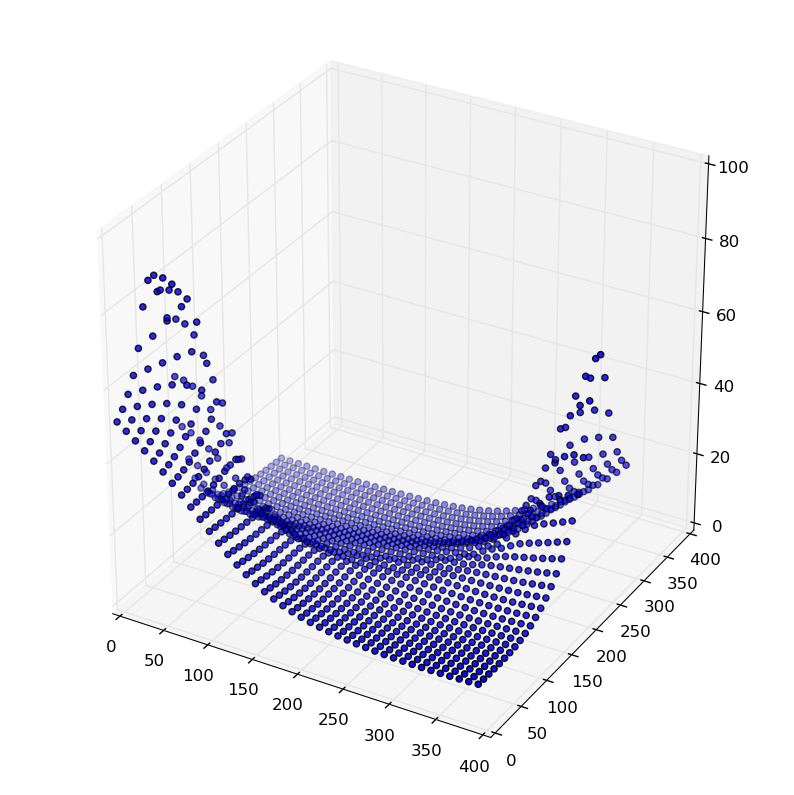} &
	\includegraphics{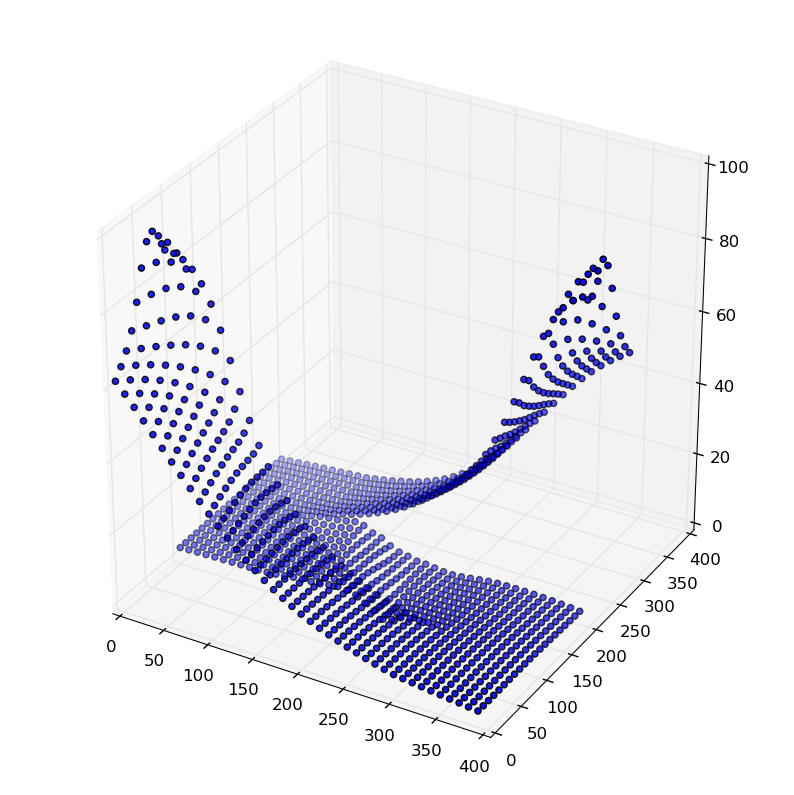}
\end{tabularx}
\caption{Left: MSVE with and without the oracle embedding. Estimated value function without the oracle (middle) and with the oracle (right). Top: MC, Bottom: TD. Note the sharp discontinuities in the walls regions, absence of leakage, and lower MSVE, especially in the TD case.}
\label{fig:oracle_vs_baseline}
\end{figure}

Figure (\ref{fig:oracle_embedding_angles}) shows the transformation of the input space that we defined for the S-shaped map layout. As we can see in the results from Figure (\ref{fig:oracle_vs_baseline}), the solution found by TD becomes indeed much sharper and the MSVE is greatly reduced.

This is encouraging, but we have assumed privileged information to create the transformation. What if we could do it in an unsupervised manner, purely from the existing trajectories data?

\subsection{Time-proximity Embeddings}

There have been several recent works \citep{sermanet2017time, savinov2018semi, aytar2018playing} around the idea of predicting whether two states or observations, are temporally close to each other, based on a dataset of trajectories generated by a policy. To test whether these could be used to induce a good intermediate representation, we have implemented such a temporal proximity classifier. We defined multiple bins for different ranges of (discounted) time intervals into the future and trained it with self-supervised information from the trajectories.
More specifically, the input of the classifier is a pair of states from the same episode $\Delta t$ steps apart, where $\Delta t$ is sampled according to a geometric distribution of discount $\gamma$ (the same as in the MDP), sampling more often frames close in time. The $K$ bins are created such that they are equally likely, i.e. the upper bound of bin \#k is then given by:
\begin{displaymath}
T_k = \frac{\ln(1 - p)}{\ln \gamma} \quad \text{with} \quad p = \frac{k}{K}
\end{displaymath}
The last bin is reserved for states that are very far apart in time. To increase the diversity in training, we also sample pairs of states coming each from a different episode. Two states sampled this way are assumed to be far apart in time with high probability, thus we label them as belonging to the last bin \#$K$.
The estimated distribution of the time bins for a pair of states $s_1$ and $s_2$ is given by $g_{\bf w_c}(f(s_1, {\bf w_e}), f(s_2, {\bf w_e}))$ where both $f$ and $g$ are functions parameterised with a neural network with weights ${\bf w_e}$ and ${\bf w_c}$, respectively. After training is complete we only keep $f(s, {\bf w_e})$, which defines the embedding function. 

\begin{figure*}[!h]
\begin{center}
\setkeys{Gin}{width=1.0\linewidth}
\begin{tabularx}{1.0\linewidth}{X@{\hskip .7in}X@{\hskip .7in}X}
\subfloat[No embedding]{\includegraphics{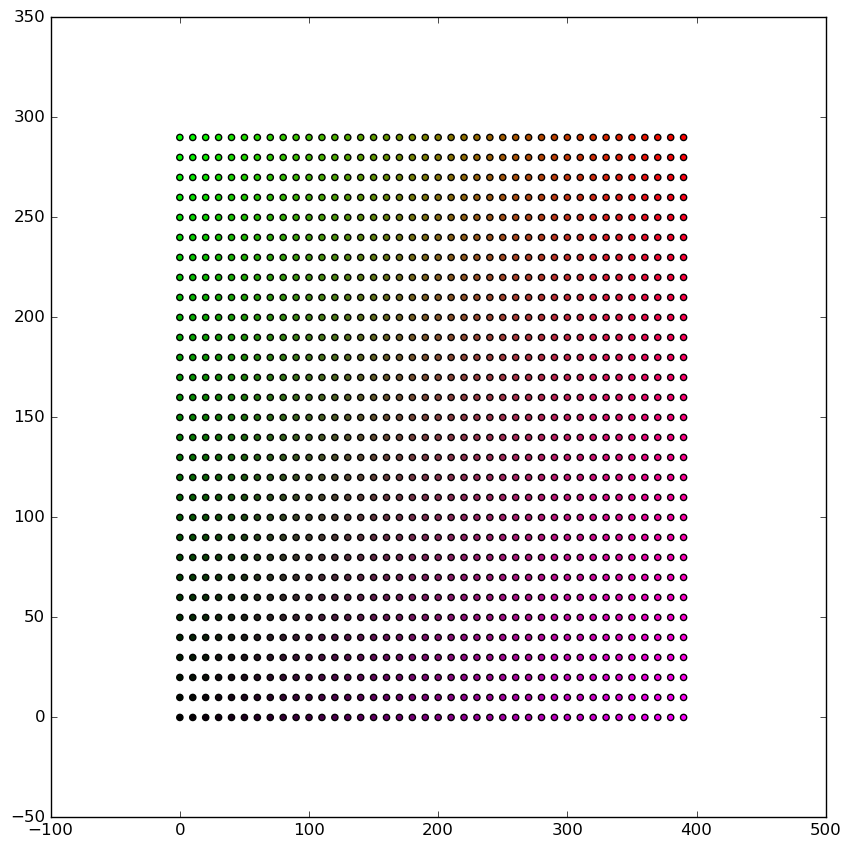}} &
\subfloat[Successor features]{\includegraphics{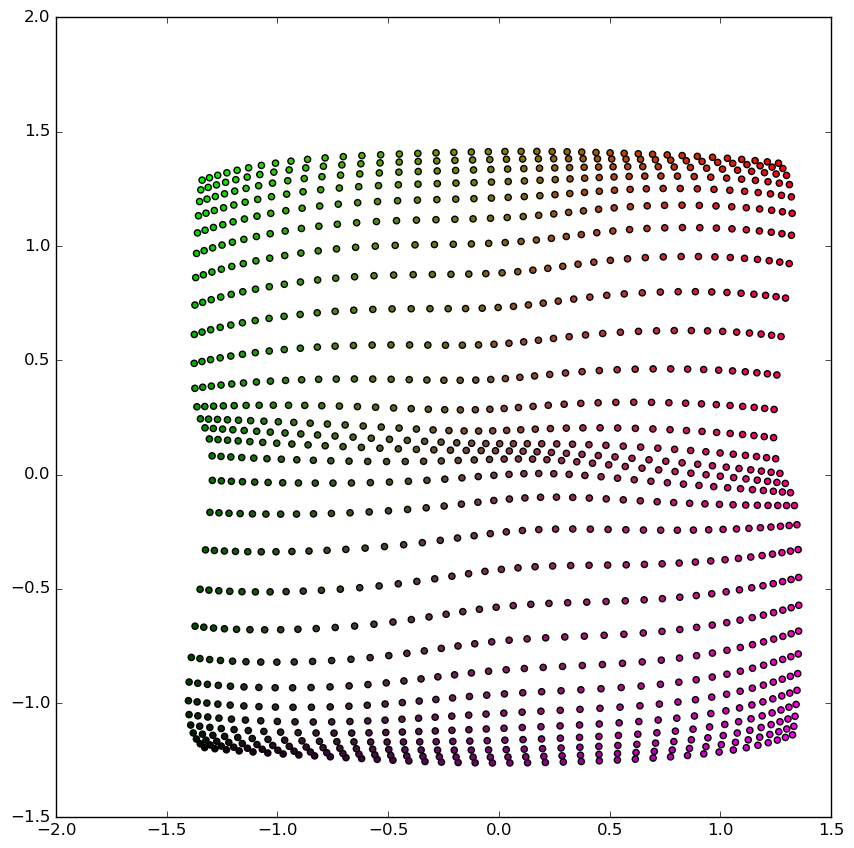}} &
\subfloat[Time-proximity]{\includegraphics{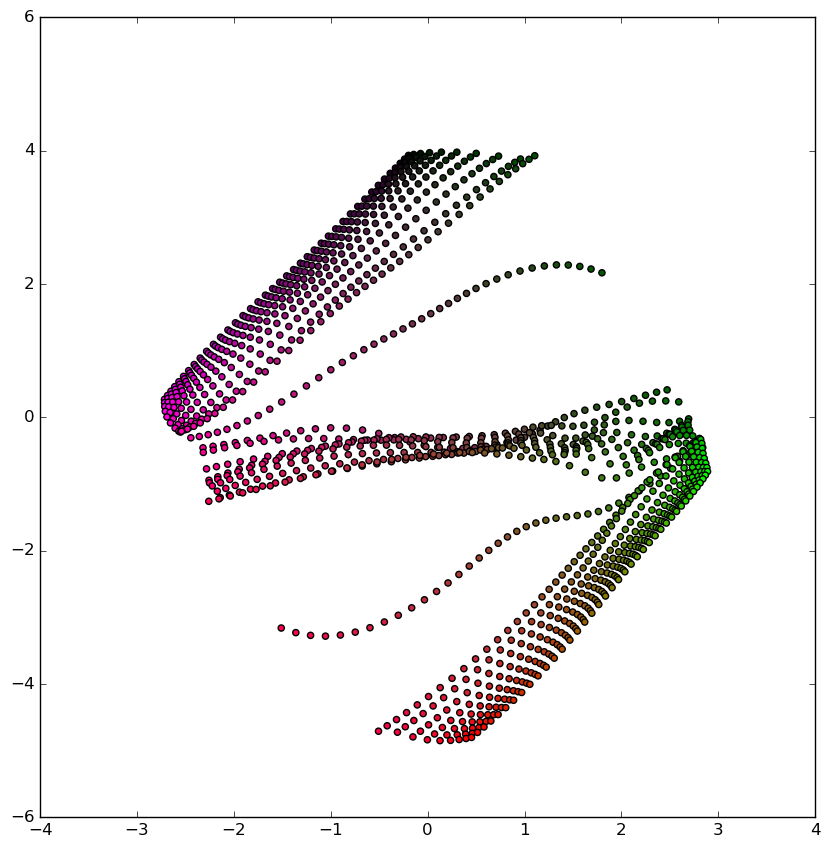}} \\
\end{tabularx}
\caption{Warping of the original 2-dimensional space on the S-shaped layout (map 1), using the embedding networks trained with the successor features or the time-proximity classifier.}
\label{fig:embedding_sf_tcn}
\end{center}
\end{figure*}

\begin{figure*}[!h]
\begin{center}
\setkeys{Gin}{width=1.0\linewidth}
\begin{tabularx}{1.0\linewidth}{XXX}
\subfloat[TD on map 1]{\includegraphics{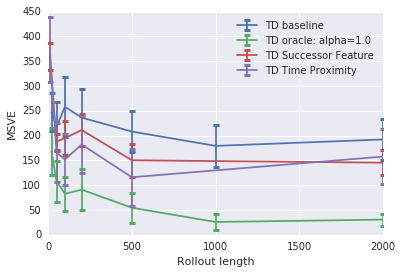}} &
\subfloat[TD on map 2]{\includegraphics{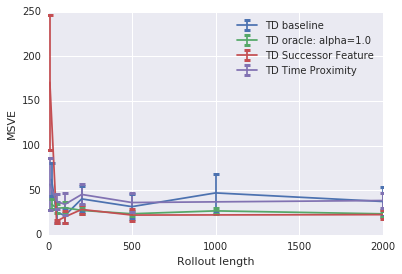}} &
\subfloat[TD on map 3]{\includegraphics{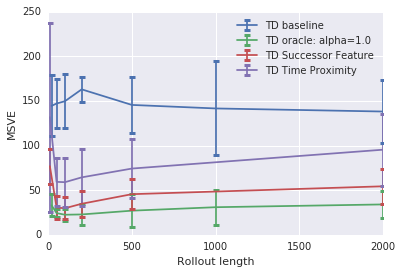}} \\
\end{tabularx}
\caption{MSVE of value function estimates learnt on top of the time proximity and the successor feature embeddings, with varying amounts of training data available (changing trajectory lengths). The hand-designed oracle embedding curves are also shown, for comparison.}
\label{fig:sf_vs_tcn_vs_baseline}
\end{center}
\end{figure*}

The results in Figure (\ref{fig:sf_vs_tcn_vs_baseline}) show that by taking the intermediate layer representation from the time proximity classifier, and then performing TD to estimate the value function on top of it, the overall Mean Square Value Error tends to be lower.
Naturally, the improvement is not as drastic as with the Oracle transformation, but it achieves a significant gain without privileged knowledge. We now investigate an alternative way of learning an intermediate representation, relying on a well-established concept from the RL literature.

\subsection{Successor Features}

The successor features concept was introduced in \cite{kulkarni2016deep} and \cite{barreto2017successor} as an extension of the successor representation \citep{dayan1993improving} for a continuous state space.
The successor representation $\Psi(s,s')$ between two states $s$ and $s'$ is defined in the tabular case as the expected discounted time to reach $s'$ from $s$ when following policy $\pi$:
\begin{equation*}
	\Psi_{\pi, s'}(s) = \E_{\pi}\left[ \sum_{t} \gamma^t \mathbbm{1}_{S_t = s'} | S_0 = s \right]
\end{equation*}
The successor representation is suitable as an intermediate embedding, as the value function can be obtained directly from this embedding through a linear operation:
\begin{equation*}
	v_{\pi}(s) = \sum_{s' \in S} \Psi_{\pi, s'}(s) r(s')
\end{equation*}

In the continuous setting, the infinite family of functions $\mathbbm{1}_{s'}$ is not suitable anymore, all the expectations would be 0. The successor features use a finite set of functions $\phi_k$ instead:
\begin{equation*}
	\Psi_{\pi, k}(s) = \E_{\pi}\left[ \sum_{t} \gamma^t \phi_k(S_t) | S_0 = s \right] .
\end{equation*}

\begin{figure}[ht]
\setkeys{Gin}{width=\linewidth}
\begin{tabularx}{\columnwidth}{XXX}

\subfloat[Map 1]{\includegraphics{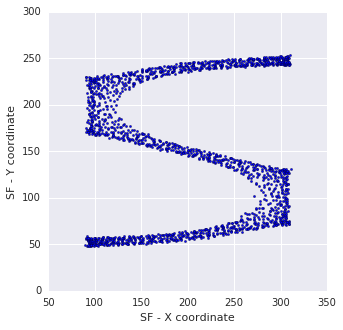}} &
\subfloat[Map 2]{\includegraphics{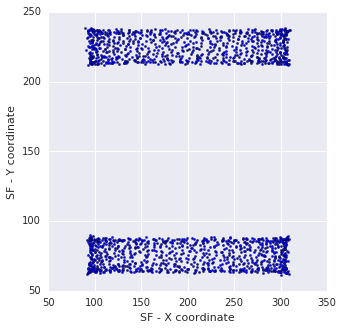}} &
\subfloat[Map 3]{\includegraphics{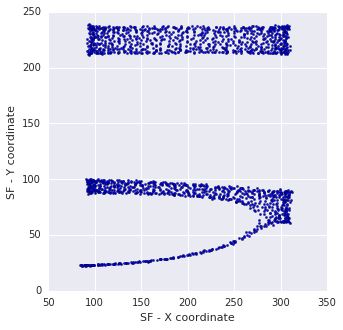}}
\end{tabularx}
\caption{Successor Features targets with $\gamma = 0.99$}
\label{fig:successor_feature_ground_truth}
\end{figure}

In the labyrinth toy environment, we use 2 features: $\phi_0(s) = x$ and $\phi_1(s) = y$. The successor features now define the average discounted position of the agent in the labyrinth. An example is given by Figure (\ref{fig:successor_feature_ground_truth}): one can notice that an initial position that is close to the wall can only get further away from the wall along the trajectories. Hence the successor features of two points in opposite sides of a wall will be significantly different.

As we can see from Figure (\ref{fig:sf_vs_tcn_vs_baseline}), the successor representation for states allows TD to find a  state-value function with significantly lower Mean-Squared Value Error.
In our setup it is \textit{crucial} that the successor features are first learned with Monte Carlo targets and \textit{then} used by Temporal Difference to learn the value function. If we would learn the successor features with TD, we would immediately face the leakage propagation problem which we intend to avoid.

\section{Experimental details}

Below we show the choices of hyper-parameters related to the environments, data generation, neural network architecture and the training procedure. \\

\begin{tabular}{|l|l|}
  \hline
  \multicolumn{2}{|l|}{\textbf{Data}} \\
  \hline \hline
  Environments size (W x H) & 400 x 300 \\
  Environment discount factor & 0.99 \\
  Environment reward areas radius & 10 \\
  Environment reward  & +30 \\
  Agent number of actions & 360 (angles) \\
  Agent step size & 20 \\
  Trajectory max length & 2000 \\
  Number of training trajectories & 100 \\
  \hline
\end{tabular} \\ 
\\
\begin{center}
\begin{tabular}{|l|l|}
  \hline
  \multicolumn{2}{|l|}{\textbf{Multi-layer Perceptron architecture}}  \\
  \hline \hline
  Embedding layers sizes & [20, 20, 20, 2] \\
  Value layers sizes & [30, 30, 1] \\
  Non-linearities & tanh \\
  \hline
  \hline
  \multicolumn{2}{|l|}{\textbf{Training hyper-parameters}}  \\
  \hline
  \hline
  Mini-batch size & 32 \\
  Optimiser algorithm & Adam \\
  Learning rate & 0.001 \\
  Beta1 & 0.9 \\
  Beta2 & 0.999 \\
  Epsilon & 1e-08 \\
  Embedding training steps & 40000 \\
  Value training steps & 40000 \\
  \hline
\end{tabular}
\end{center}

Note that we fixed the output of the embedding network to always be 2-dimensional, so that we could easily visualise it. This is also the case for the hand-designed embedding.
To generate the error bars in our plots, we run each configuration with 20 different random seeds.

\section{Related Work}

Reinforcement Learning books by \cite{sutton1998introduction, sutton2018introduction} and \cite{szepesvari2010algorithms}, cover the general ideas behind on-policy evaluation via TD and MC, with and without function approximation. In the following sub-sections we point to more specific research articles in the literature.

\subsection{Temporal Difference Learning}


TD Learning with function approximation has been analysed by \cite{tsitsiklis1997analysis}, with convergence results for the on-policy linear case, and showcasing divergence examples with off-policy data. \cite{baird1995residual} introduces convergent algorithms that minimise the Bellman residual directly, but they can be slow. More concerning, if the environment is stochastic, they would require two independent transitions from the same state. This is often unrealistic, most notably when dealing with very large or continuous state spaces. \cite{baird1995residual} also describes how a value function with some parameter sharing, can propagate information ``the wrong direction", by not respecting temporal causality.
\cite{singh1995reinforcement} deals with aggregation of states, in a soft-clustering setup, which does not cover other function approximators of interest, like neural networks.

\citet{sutton2009fast, sutton2009convergent} introduce two modified algorithms for TD with linear function approximation that provably converge in the off-policy setting. \citet{adam2016investigating} provide a comparison of linear TD methods and recommendations on which variant to pick depending on the problem constraints. \citet{bhatnagar2009convergent} introduces a version of TD learning that converges under the off-policy setting with a large class of smooth function approximators, such as neural networks. However, the solutions found might have poor Mean-squared Value Error (MSVE).

In a recent paper, \cite{ollivier2018approximate} shows that for the particular case of \textit{reversible policies}, TD does gradient descent of a loss function that includes a term with the Dirichlet norm $\lVert \hat{v} - v \rVert^{2}_{Dir}$. This formulation provides a more natural way to think about the fixed point solution for TD, for reversible policies, as a tension between two competing objectives.

Great part of the RL literature focus on online learning, however, there is also a set of works devoted to the offline, or \textit{batch} case. In this setup, all the data is assumed to be pre-collected and available at training time.
In analogy with the well established least-squares regression, \cite{bradtke1996linear} introduces the first linear least-squares algorithms for temporal-difference learning (LSTD). This has the advantage of making optimal use of all the training data available (as opposed to some ``forgetting" in online gradient methods), however it has higher memory resources requirements, and it is limited to linear function approximation. \cite{boyan1999least} provides a simpler derivation of LSTD and extends that work by generalising from $\lambda = 1$ to arbitrary lambda values.



\cite{mann2016adaptive} deals with a setup very similar to ours: accurate on-policy evaluation from a dataset of trajectories. They suggest tuning the $\lambda$ parameter of TD($\lambda$) with leave-one-trajectory-out cross-validation, in order to minimise the final MSVE. However, the suggested algorithm can only do it efficiently for linear function approximation.

\subsection{Successor Representations}

The concept of \textit{Successor Representation} was originally introduced by \citep{dayan1993improving}. That influential paper, already sets the main intuition that we pursue in the second half of this work: ``\textit{For static tasks, generalisation is typically sought by awarding similar representations to states that are nearby in some space. This concept extends to tasks involving predictions over time, except that adjacency is defined in terms of similarity of the future course of the behaviour of a dynamic system.}". In our work though, we focus on the setup of continuous state spaces and neural networks trained with gradient methods, rather than finite state Markov Decision Processes (MDP) and linear function approximators. We also do not assume knowledge of environment dynamics and/or the policy specification, restricting ourselves to a dataset of trajectories, as in Batch RL. \cite{dayan1993improving}, who experimented with a control task, hinted that some of the problems of having a policy-dependent representation would not be an issue if one would care only about pure estimation tasks. It turns out this is our problem of interest, as we just want to be more accurate at evaluating the current policy, rather than finding a better one.

Extensions to continuous state spaces, named \textit{Deep Successor Representation} or \textit{Successor Features} were introduced by \citep{kulkarni2016deep} and \citep{barreto2017successor}, respectively. \citep{kulkarni2016deep} gives a constructive way of learning these representations, by using auxiliary losses that shape the representation to keep the necessary information to predict the immediate reward and next observation. Their aim is to solve control/navigation tasks and the automatic extraction of sub-goals, often from raw pixels of the observations. \cite{barreto2017successor} focuses more on using successor features for transfer learning, in setups where the reward function changes but the environment dynamics remains the same.

All these works highlight the importance of the factorisation of the value function into a dot product of the successor features and a weight vector, and mention that the representation can be learned by any RL algorithm (e.g. Monte-Carlo or TD itself). In our work, though, we put less emphasis on the factorisation and only use successor features as a way to distinguish deceptively similar states. In addition, we use it as a mechanism to avoid some of the problems of using TD with neural networks. In this setup, we recommend \textit{against} using TD to learn the Successor Features. We recommend first learning the successor features with Monte Carlo regression and \textit{then} use them to learn the value function on top with Temporal Difference, so as to avoid the problems of leakage propagation.






Proto-value functions by \cite{mahadevan2005proto, mahadevan2007proto} also try to tackle the problem of having a better representation for learning value functions. The original paper provided inspiration for our representation learning approaches: "\textit{... learned not from rewards, but instead from analysing the topology of the state space}", and the experimental scenarios we test: \textit{``states close in Euclidean distance may be far apart on the manifold (e.g, two states on opposite sides of a wall).}". It has been shown that proto-value functions and the successor representation in discrete state spaces are fundamentally the same thing, and that \textit{successor features} are a way of extending the concept to continuous state spaces. We therefore focused more on the use of successor features.

\subsection{Deep RL and the deadly triad}

\cite{sutton2018introduction} explain that three ``deadly" components are needed to have divergence: \textit{function approximation}, \textit{bootstrapping} and \textit{off-policy} data. Great progress has been made at improving the stability of Deep Reinforcement Learning with techniques like Target Networks and Experience Replays  \citep{mnih2015human, schaul2015prioritized}, but these focus more on the setup of \textit{off-policy} learning for control tasks.

In our work, we do not have the problem of off-policy data, therefore are protected from divergence. Our focus is  not on whether TD converges or not, but what solution it converges to, and how good it is. We studied a particular pathology, that we called leakage propagation, under a somewhat restricted scenario of random walks in navigation tasks. However, we believe it illustrates well the problems that should also be visible in the broader setup.

\section{Conclusions and Future Work}

We described the \textit{leakage propagation} problem that can happen in on-policy evaluation with Temporal-Difference learning and neural networks, and tried to deepen our understanding of it. We showed visual empirical evidence to develop intuition about the problem, in environments with sharp discontinuities in seemingly nearby states. We then gave a formal understanding of the phenomenon in a simple reversible MDP, by analytically finding the minimum of the Dirichlet and the Euclidean norms mixture.
We showed that privileged-knowledge representations can mitigate the problem and bring significant estimation accuracy gains; we then suggested ways of using unsupervised learning to get partway there. The additional unsupervised losses are simple to implement, and make use of topological information about the trajectories, that neither Monte-Carlo nor Temporal-Difference learning explicitly exploit. These heuristics worked reasonably well in the simple two-dimensional navigation environments. However, we make no claims that these are directly extensible to more complex problems with high-dimensional inputs, as one would have to ensure that no important information is discarded in the process of learning the embedding. This requires further research.
Another line of future research would be to devise a criterion to decide, per state, whether doing a TD bootstrap update is "trustworthy" or whether it would be better to rely on longer n-step updates. This would be an interesting alternative to the representation learning approach.

\subsubsection*{Acknowledgements}

We would like to thank Cosmin Paduraru, Will Dabney, Tom Schaul and Hado van Hasselt  for helpful and insightful discussions.

\bibliography{extended_version}
\bibliographystyle{plainnat}




\end{document}